\documentclass[10pt,twocolumn,letterpaper]{article}

\usepackage{float}
\usepackage{placeins}
\usepackage{multirow}
\usepackage{tikz}
\usepackage{pgfplots}
\usepackage{array, makecell}
\usepackage{mwe}

\usepackage{algorithm}
\usepackage{algorithmic}
\usepackage{bm}
\usepackage[table]{xcolor}
\usepackage{graphicx}
\usepackage{makecell}
\usepackage{amsmath}
\usepackage{float}
\usepackage{hhline}
\usepackage{geometry}
\usepackage{booktabs}
\usepackage{arydshln}
\usepackage{amsfonts}
\usepackage{multirow}
\usepackage{subcaption}
\usepackage[table]{xcolor}  
\definecolor{bestColor}{RGB}{255, 0, 0}    
\definecolor{secondBestColor}{RGB}{0, 0, 255} 
\definecolor{thirdBestColor}{RGB}{240,0, 240}  
\definecolor{qBestColor}{RGB}{0,215, 0}  

\definecolor{secondBestColor}{RGB}{0, 0, 255} 
\definecolor{thirdBestColor}{RGB}{240,0, 240}  
\definecolor{qBestColor}{RGB}{0,215, 0}  

\definecolor{cnn_color}{RGB}{240,248,255}  
\definecolor{incompression_color}{RGB}{255,250,240}  
\definecolor{crosscompression_color}{RGB}{240,255,240}  
\definecolor{input_color}{RGB}{255,240,240}      
\definecolor{mamba_color}{RGB}{235,235,225}  

\definecolor{deepred}{RGB}{133,0,0}    
\definecolor{lightred}{RGB}{255,102,102}  
\definecolor{deepgreen}{RGB}{0,133,0}  

\definecolor{bestcrossmanupulation_color}{RGB}{255,240,240} 
\definecolor{secondcrossmanupulation_color}{RGB}{240,240,255} 
\definecolor{thirdcrossmanupulation_color}{RGB}{240,255,240} 
\definecolor{background_color}{RGB}{255,255,255}

\usepackage{pifont}
\usepackage{cvpr}              

\usepackage{graphicx}

\usepackage{tikz}
\usetikzlibrary{arrows.meta,positioning,calc,fit,shapes,backgrounds}

\definecolor{deepred}{RGB}{200,0,0}
\definecolor{deeporange}{RGB}{230,120,20}
\definecolor{deepgold}{RGB}{184,134,11}
\definecolor{deepgreen}{RGB}{0,128,0}
\definecolor{deepblue}{RGB}{0,0,180}
\definecolor{deepviolet}{RGB}{138,43,226}

\definecolor{cvprblue}{rgb}{0.21,0.49,0.74}
\usepackage[pagebackref,breaklinks,colorlinks,
            citecolor=cvprblue,
            linkcolor=cvprblue,
            urlcolor=cvprblue]{hyperref}
\renewcommand*{\backref}[1]{}
\renewcommand*{\backrefalt}[4]{%
  \ifcase #1 \or {\color{cvprblue}#2}\else {\color{cvprblue}#2}\fi}

\def\paperID{9} 
\def\confName{CVPR}
\def\confYear{2026}

\makeatletter
\def\blfootnote{\xdef\@thefnmark{}\@footnotetext}
\makeatother

\title{Cross-modal Affinity-aligned Multimodal Learning Analytics for Predicting Student Collaboration Satisfaction in Game-Based Learning}

\author{Wen-Hsin Tsai$^{1}$ \quad Chia-Ming Lee$^{2,3}$ \quad Yuk-Ying Tung$^1$\\
{$^1$Institute of Education, National Cheng Kung University} \\{$^2$Institute of Intelligent System, National Yang Ming Chiao Tung University}\\ {$^3$Department of Computer Science, University at Albany, State University of New York}}
\begin{document}
\maketitle

\begin{abstract}
Collaborative game-based learning environments offer rich opportunities 
for small-group knowledge construction, yet automatically predicting 
student collaboration satisfaction remains challenging. A critical barrier 
is modality degradation: in educational deployments, individual modalities 
such as eye gaze exhibit inconsistent informativeness across student 
cohorts, causing implicit attention-based fusion to produce brittle 
multimodal representations. We propose the Affinity-Aligned Multimodal 
Learning Analytics (AAMLA) framework, whose core contribution is the 
Cross-modal Affinity-guided Modality Alignment (CAMA) module, which 
explicitly models inter-modal relationships via affinity matrices and 
enforces cross-modal consistency through contrastive learning, enabling 
adaptive suppression of uninformative modalities without discarding them. 
AAMLA further applies modality-specific projection layers to map 
heterogeneous features — facial action units, head pose, eye gaze, and 
interaction trace logs — into a unified semantic space prior to alignment. 
Experiments on 50 middle school students in the EcoJourneys collaborative 
learning environment demonstrate consistent improvements over unimodal 
baselines and prior cross-attention approaches under standard and modality 
degradation conditions, with SHAP and t-SNE analyses confirming that CAMA 
produces robust, interpretable cross-modal representations for student 
collaboration modeling.
\end{abstract}
    
\section{Introduction}

Collaborative game-based learning environments provide students with
immersive opportunities for small-group problem solving, knowledge
construction, and collaborative inquiry~\cite{zambrano2019cognitive,
zhu2012student,fonseca2023gamification,hava2020gifted}. Within these
environments, student satisfaction with collaborative experiences
plays a pivotal role in shaping teamwork dynamics, sustaining motivation,
and ultimately influencing learning outcomes~\cite{zambrano2019cognitive}.
Accurately gauging collaboration satisfaction can offer actionable insights
for instructors and intelligent tutoring systems to provide timely,
targeted interventions that support productive group interactions. However,
existing work predominantly relies on post-hoc survey
analysis~\cite{al2013evaluating,lee2011examining,so2008student}, leaving
automated assessment methods relatively underexplored.

Multimodal behavioral signals, including facial action units, head pose,
eye gaze, and interaction trace logs, offer rich and complementary cues
for understanding student collaboration dynamics~\cite{schneider2014collaboration,
dich2018physiological}. However, a fundamental challenge in educational
deployments is that \textbf{individual modalities are frequently unreliable in
practice}: sensor noise, occlusion, and student behavioral variation cause
modality quality to fluctuate across students and sessions~\cite{ma2021smil,
wang2023multi}. This problem is especially pronounced for gaze features,
which exhibit inconsistent informativeness across student cohorts~\cite{
guo2020deep,harris2022looking}, a form of modality degradation that
cannot be addressed by simply discarding the affected modality, as it may
still carry useful signals in other contexts. Despite these challenges,
existing multimodal approaches in learning analytics rely on implicit
cross-attention fusion mechanisms~\cite{li2021selfdoc} that assume all
modalities contribute equally informative signals, and lack explicit
mechanisms to handle modality degradation gracefully. When a weak or noisy
modality dominates attention, cascading errors propagate through the entire
fusion pipeline, degrading prediction robustness across student cohorts.

To address these limitations, we propose the \textbf{Affinity-Aligned
Multimodal Learning Analytics (AAMLA)} framework for predicting student
collaboration satisfaction in collaborative game-based learning. The core
contribution of AAMLA is the \textbf{Cross-modal Affinity-guided Modality
Alignment (CAMA)} module, which explicitly models inter-modal relationships
through affinity matrices and enforces cross-modal consistency via
contrastive learning~\cite{chen2021intriguing,wang2023connecting}. Unlike
implicit fusion, CAMA learns to adaptively suppress uninformative or
degraded modalities, such as inconsistent gaze signals, without
discarding them entirely, ensuring stable multimodal representations even
under severe modality degradation. We further apply modality-specific
linear projection layers to map heterogeneous features,
including facial action units~\cite{baltrusaitis2018openface}, head pose, gaze, and
BERT-based trace log embeddings~\cite{li2021selfdoc}, into a unified
semantic space prior to alignment, enabling direct cross-modal comparison.

Our research addresses the following questions:
\begin{itemize}
    \item \textbf{RQ1:} What are the individual modalities that contribute
    most significantly to predicting student collaboration
    satisfaction~\cite{bradford2023automatic,stewart2021multimodal,
    starr2018toward}?
    \item \textbf{RQ2:} How does CAMA improve robustness over implicit
    cross-attention fusion under modality degradation conditions including
    gaze dropout, feature perturbation, and full modality absence?
    \item \textbf{RQ3:} What modality combinations are most effective, and
    what do affinity matrix visualizations and t-SNE distributions reveal
    about cross-modal alignment quality under degradation?
\end{itemize}

We conduct experiments on multimodal data collected from 50 middle school
students interacting with the EcoJourneys collaborative game-based learning
environment~\cite{carpenter2020detecting}. Results demonstrate that AAMLA
consistently outperforms unimodal baselines and the prior cross-attention
approach~\cite{EDM19} across all modality combinations, and maintains
robust performance under three categories of modality degradation.
t-SNE visualizations~\cite{van2008visualizing} and affinity matrix
analyses confirm that CAMA produces semantically coherent cross-modal
representations, offering interpretable insights into the multimodal
behavioral patterns most predictive of student collaboration satisfaction.
\section{Related Work}
\label{sec:related}

\noindent\textbf{Collaborative Game-Based Learning.} Collaborative game-based learning environments offer significant promise in fostering collaborative skills and providing insights into student dynamics~\cite{carpenter2020detecting,li2023socially,liang2023mandatory,chen2022bibliometric}. Digital games serve as a conduit to explore the relationship between learning outcomes and collaborative gameplay~\cite{baek2020comparing,puga2022game}. However, a systematic framework elucidating the elements of collaborative learning within these environments remains elusive~\cite{wang2021systematic}, while the pivotal importance of student satisfaction and motivation in educational tool design has been consistently highlighted~\cite{fonseca2023gamification}.

\noindent\textbf{Student Satisfaction.} Prior work has investigated the relationship between student satisfaction and learning processes across diverse settings~\cite{al2013evaluating,lee2011examining,so2008student,yunusa2021scoping}. Collaborative engagement and instructional quality emerge as pivotal determinants of satisfaction, which is consistently associated with higher learning outcomes~\cite{yu2021effect}. However, existing research predominantly relies on post-hoc surveys, overlooking automated assessment methodologies. The closest prior work~\cite{ma2022investigating} predicts peer satisfaction in dyadic settings from multimodal cues, yet extending this to non-dyadic small groups remains an open challenge.

\noindent\textbf{Multimodal Learning Analytics.} Multimodal learning analytics provides rich indicators of collaborative engagement~\cite{starr2018toward,ma2022detecting,bradford2023automatic,stewart2021multimodal}, with prior work demonstrating effectiveness in predicting affective states during collaborative activities~\cite{daoudi2020edm}, modeling cognitive load from physiological signals~\cite{cai2022modeling}, and identifying collaboration quality from speech and non-verbal cues~\cite{praharaj2022towards}. Griffith et al.~\cite{griffith2023investigating} further demonstrate how co-creative dialogue states influence partner satisfaction, while temporal integration of multimodal features has been shown to improve prediction of learning gains~\cite{olsen2020temporal}. Despite these advances, existing frameworks rely on implicit fusion mechanisms that treat all modalities equally, without accounting for the varying informativeness of individual modalities across different student cohorts and interaction contexts. The exploration of collaboration satisfaction prediction in game-based learning environments, particularly under such modality imbalance, remains underexplored.

\noindent\textbf{Modality Degradation and Robustness.} A fundamental challenge in multimodal learning is that individual modalities may be unavailable, corrupted, or simply uninformative in practice~\cite{ma2021smil,wang2023multi}. In educational settings, this problem is especially pronounced: sensor noise, occlusion, and student behavioral variation can render certain modalities unreliable across sessions or cohorts. Prior work has explored missing modality scenarios through shared-specific feature modeling~\cite{wang2023multi} and Bayesian approaches for severely missing modalities~\cite{ma2021smil}, while test-time adaptation methods have been proposed to handle modality reliability bias at inference~\cite{yang2024testtime}. In the context of deepfake detection, Le and Woo~\cite{QAD} demonstrate that quality-agnostic learning can improve robustness against compression-induced feature degradation, and UMCL~\cite{umcl} further shows that generating complementary modalities from a single source mitigates unequal modality degradation. However, these approaches are not designed for the educational domain, where modality degradation is behavioral rather than compression-induced. In our setting, gaze features exhibit inconsistent informativeness across student cohorts~\cite{guo2020deep,harris2022looking}, motivating an explicit mechanism to suppress uninformative modalities dynamically rather than discarding them entirely.

\noindent\textbf{Feature Alignment in Multimodal Learning.} Feature alignment is fundamental to robust multimodal fusion, ensuring that modality-specific representations occupy a coherent shared semantic space~\cite{wang2024structureclip,wang2024lammprompt,wang2024mmap}. Contrastive learning approaches reduce modality gaps through explicit cross-modal regularization~\cite{chen2021intriguing,liang2022mind,wang2023connecting}, while affinity-based methods model global and local inter-modal relationships to improve interpretability and robustness~\cite{lee2021looking}. Recent work in multimodal deepfake detection~\cite{umcl} demonstrates that explicit alignment of self-generated modalities substantially outperforms implicit attention-based fusion under degraded conditions. Nevertheless, existing alignment methods largely assume that all modalities contribute meaningful signal — an assumption that fails when weak modalities introduce noise into the fusion process. Our proposed CAMA strategy extends affinity-based alignment to explicitly down-weight uninformative modalities, ensuring stable cross-modal coordination even under behavioral modality degradation in collaborative learning settings.
\section{Methodology}
\label{sec:method}

\subsection{Framework Overview}

\begin{figure*}[h]
    \centering
    \includegraphics[width=1.0\textwidth]{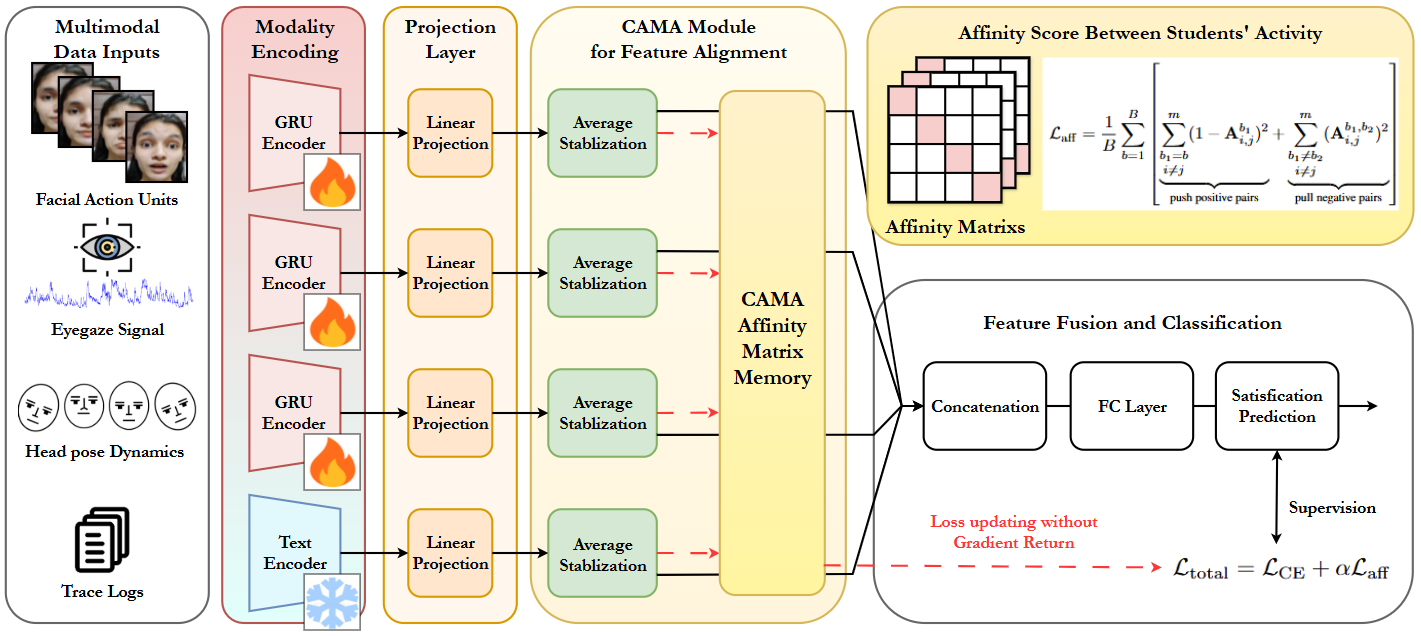}
    \caption{\textbf{Overview of the proposed AAMLA framework.} Four 
    modality streams (facial action units, head pose, eye gaze, trace logs) 
    are encoded by modality-specific encoders and projected into a unified 
    $d=128$ semantic space. The CAMA module explicitly models inter-modal 
    relationships via affinity matrices and contrastive loss 
    $\mathcal{L}_\mathrm{aff}$, suppressing uninformative modalities. 
    Aligned embeddings are classified by a FC head optimized with 
    $\mathcal{L}_\mathrm{CE}$, producing four-class collaboration 
    satisfaction predictions.}
    \label{fig:framework}
\end{figure*}
In this section, we present the AAMLA framework for collaboration 
satisfaction prediction. As illustrated in Fig.~\ref{fig:framework}, AAMLA 
processes four complementary modality streams from student interaction 
data, including facial action units (AU), head pose, eye gaze, and trace logs, 
through modality-specific encoders. The resulting embeddings are unified 
via learnable projection layers and aligned by the proposed \textbf{Cross-modal 
Affinity-guided Modality Alignment (CAMA)} module, which explicitly models 
inter-modal relationships through affinity matrices to suppress uninformative 
modalities and ensure semantic consistency across heterogeneous feature spaces.

\subsection{Multimodal Feature Encoding}

Our framework operates on four complementary modalities extracted from 
student interactions within the EcoJourneys learning 
environment~\cite{carpenter2020detecting}. Each modality captures distinct 
aspects of collaborative behavior that contribute uniquely to satisfaction 
prediction.

\noindent\textbf{Facial Action Units.}
Facial action units (AUs) capture the contractions and relaxations of 
facial muscles, providing direct cues about students' emotional states and 
engagement during collaboration~\cite{mcdaniel2017facial,borges2019classifying}. 
We extract 17 AU intensity values per frame using OpenFace 
2.0~\cite{baltrusaitis2018openface}, producing a sequence 
$\mathbf{X}_{\mathrm{AU}} \in \mathbb{R}^{T \times 17}$ over $T$ 
timesteps. The AU encoder processes this sequence through a GRU:
\begin{equation}
    \mathbf{e}_{\mathrm{AU}} = \Phi_{\mathrm{AU}}(\mathbf{X}_{\mathrm{AU}}),
    \label{eq:au-encoder}
\end{equation}
where $\Phi_{\mathrm{AU}}$ represents the GRU-based AU encoder and 
$\mathbf{e}_{\mathrm{AU}} \in \mathbb{R}^{1 \times d_{\mathrm{AU}}}$ is 
the resulting temporal embedding.

\noindent\textbf{Head Pose.}
Head pose information captures students' head location and orientation 
relative to the camera, providing cues about attentiveness and engagement 
during collaborative activities~\cite{sumer2021multimodal,chejara2023exploring}. 
We extract 6 pose features per frame (3D translation and rotation 
coordinates) using OpenFace 2.0~\cite{baltrusaitis2018openface}, producing 
$\mathbf{X}_{\mathrm{pose}} \in \mathbb{R}^{T \times 6}$:
\begin{equation}
    \mathbf{e}_{\mathrm{pose}} = \Phi_{\mathrm{pose}}(\mathbf{X}_{\mathrm{pose}}),
    \label{eq:pose-encoder}
\end{equation}
where $\Phi_{\mathrm{pose}}$ represents the GRU-based pose encoder.

\noindent\textbf{Eye Gaze.}
Eye gaze features detail the direction each eye is looking relative to the 
camera, capturing visual attention patterns during 
collaboration~\cite{guo2020deep,harris2022looking,sharma2020utilizing}. 
We extract 6 gaze direction features per frame (3D coordinates per eye) 
using OpenFace 2.0~\cite{baltrusaitis2018openface}, producing 
$\mathbf{X}_{\mathrm{gaze}} \in \mathbb{R}^{T \times 6}$:
\begin{equation}
    \mathbf{e}_{\mathrm{gaze}} = \Phi_{\mathrm{gaze}}(\mathbf{X}_{\mathrm{gaze}}),
    \label{eq:gaze-encoder}
\end{equation}
where $\Phi_{\mathrm{gaze}}$ represents the GRU-based gaze encoder. As 
demonstrated in prior work~\cite{ma2022investigating} and our preliminary 
analysis, \textbf{gaze features exhibit inconsistent informativeness} across student 
cohorts, motivating our explicit alignment mechanism to suppress their 
contribution dynamically when uninformative.

\noindent\textbf{Interaction Trace Logs.}
Trace logs document students' in-game actions, including NPC interactions, 
locations visited, evidence collected, and chat messages, providing rich 
contextual information about collaborative 
behavior~\cite{olsen2020temporal,griffith2023investigating}. Each trace 
event is encoded as a 768-dimensional embedding using a BERT-based sentence 
encoder fine-tuned on in-game text via unsupervised 
pre-training~\cite{li2021selfdoc}. The trace encoder projects these 
embeddings through a feedforward network:
\begin{equation}
    \mathbf{e}_{\mathrm{trace}} = \Phi_{\mathrm{trace}}(\mathbf{X}_{\mathrm{trace}}),
    \label{eq:trace-encoder}
\end{equation}
where $\Phi_{\mathrm{trace}}$ represents the feedforward trace encoder and 
$\mathbf{X}_{\mathrm{trace}} \in \mathbb{R}^{T \times 768}$.

\noindent\textbf{Feature Projection.}
After encoding, a key challenge is ensuring consistent feature 
distributions across modalities with heterogeneous dimensionalities. We 
apply modality-specific learnable linear projection layers to map all 
features into a unified $d$-dimensional semantic space:
\begin{equation}
\begin{aligned}
    \hat{\mathbf{e}}_{\mathrm{AU}} &= \mathbf{W}_{\mathrm{AU}}\mathbf{e}_{\mathrm{AU}} + \mathbf{B}_{\mathrm{AU}}, \quad
    \hat{\mathbf{e}}_{\mathrm{pose}} = \mathbf{W}_{\mathrm{pose}}\mathbf{e}_{\mathrm{pose}} + \mathbf{B}_{\mathrm{pose}},\\
    \hat{\mathbf{e}}_{\mathrm{gaze}} &= \mathbf{W}_{\mathrm{gaze}}\mathbf{e}_{\mathrm{gaze}} + \mathbf{B}_{\mathrm{gaze}}, \quad
    \hat{\mathbf{e}}_{\mathrm{trace}} = \mathbf{W}_{\mathrm{trace}}\mathbf{e}_{\mathrm{trace}} + \mathbf{B}_{\mathrm{trace}},
\end{aligned}
\label{eq:projection}
\end{equation}
where $\mathbf{W}_{\{\cdot\}}$ and $\mathbf{B}_{\{\cdot\}}$ are 
modality-specific learnable parameters, and all projected features 
$\hat{\mathbf{e}}_{\{\cdot\}} \in \mathbb{R}^{1 \times d}$ share a common 
dimensionality $d = 128$.

The projected features are concatenated to form a unified multimodal 
representation:
\begin{equation}
    \mathbf{U} = [\hat{\mathbf{e}}_{\mathrm{AU}}, \hat{\mathbf{e}}_{\mathrm{pose}}, \hat{\mathbf{e}}_{\mathrm{gaze}}, \hat{\mathbf{e}}_{\mathrm{trace}}] \in \mathbb{R}^{m \times d},
\end{equation}
where $m = 4$ represents the number of modalities. The concatenated 
features are processed through a shared fully-connected classification 
head:
\begin{equation}
    \hat{\mathbf{Y}} = \mathrm{Softmax}(\mathbf{W}_{\mathrm{FC}}\mathbf{U}),
\end{equation}
where $\mathbf{W}_{\mathrm{FC}} \in \mathbb{R}^{n_{\mathrm{out}} \times n_{\mathrm{class}}}$, 
$n_{\mathrm{class}} = 4$ corresponds to the four collaboration satisfaction 
categories, and $\hat{\mathbf{Y}}$ represents the predicted class 
probabilities. We employ cross-entropy loss for training:
\begin{equation}
\scalebox{0.88}{$\displaystyle
    \mathcal{L}_{\mathrm{CE}} = -\frac{1}{N}\sum_{i=1}^{N} \sum_{c=1}^{4} y_{i,c} \log(p_{i,c}),
    \label{eq:ce_loss}$}
\end{equation}
where $N$ is the number of training samples, $y_{i,c}$ is the ground-truth 
one-hot label, and $p_{i,c}$ is the predicted probability for class $c$.

\vspace{-2mm}
\subsection{Cross-modal Affinity-guided Alignment}

\begin{figure}[ht]
    \centering
    \includegraphics[width=0.5\textwidth]{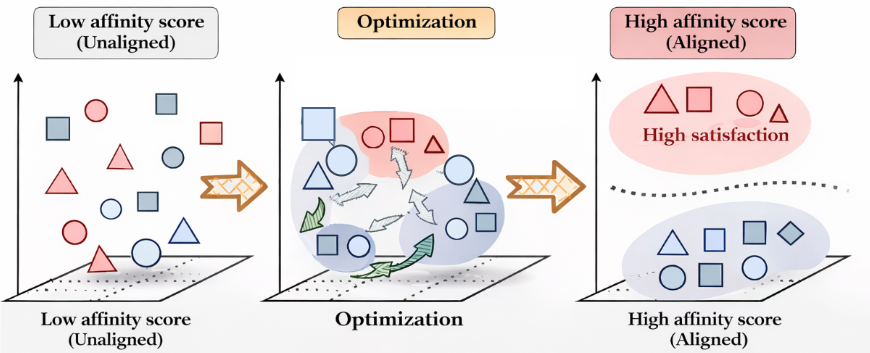}
    \caption{\textbf{Pipeline of the proposed CAMA strategy.} Different shapes denote different modalities; color denotes satisfaction class 
    (red: high; blue: low). CAMA pulls same-class modality embeddings 
    together and pushes apart different-class embeddings via affinity 
    matrices, transforming scattered unaligned features (\textit{left}) 
    into compact, semantically coherent clusters (\textit{right}) robust 
    to uninformative modalities such as gaze.}
    \label{fig:CAMA}
\vspace{-5mm}
\end{figure}

Prior cross-attention fusion mechanisms~\cite{li2021selfdoc} create 
critical vulnerabilities in multimodal student modeling: (1) 
over-concentration on dominant modalities, (2) lack of explicit 
inter-modal relationship modeling, and (3) cascading performance failures 
when dominant modalities carry noisy or inconsistent signals. In our 
setting, these vulnerabilities are particularly acute as modality 
informativeness varies significantly across student cohorts and interaction 
contexts~\cite{ma2022investigating}.

To address these limitations, we propose the CAMA strategy, which 
explicitly models inter-modal relationships through affinity matrices. As 
illustrated in Fig.~\ref{fig:CAMA}, CAMA enforces projected modality 
features into a compact and coherent shared space while ensuring \textbf{semantic 
consistency even when individual modalities degrade} (e.g., gaze occlusion, 
noisy AU detection, irregular trace sampling).

Given a batch of size $B$, batch-level modality embeddings 
$\mathbf{E}_{\mathrm{AU}}, \mathbf{E}_{\mathrm{pose}}, 
\mathbf{E}_{\mathrm{gaze}}, \mathbf{E}_{\mathrm{trace}} \in \mathbb{R}^{B 
\times d}$ are extracted using their respective encoders. To mitigate 
within-batch feature discrepancies caused by student behavioral variation, 
we compute batch-level averages across the degraded and non-degraded 
feature pairs for each modality:
\begin{equation}
\scalebox{0.88}{$\displaystyle
\begin{aligned}
    \hat{\mathbf{E}}_{\mathrm{AU}} &= \tfrac{1}{2}(\mathbf{E}_{\mathrm{AU}}^{\mathrm{orig}} + \mathbf{E}_{\mathrm{AU}}^{\mathrm{deg}}), \quad
    \hat{\mathbf{E}}_{\mathrm{pose}} = \tfrac{1}{2}(\mathbf{E}_{\mathrm{pose}}^{\mathrm{orig}} + \mathbf{E}_{\mathrm{pose}}^{\mathrm{deg}}),\\
    \hat{\mathbf{E}}_{\mathrm{gaze}} &= \tfrac{1}{2}(\mathbf{E}_{\mathrm{gaze}}^{\mathrm{orig}} + \mathbf{E}_{\mathrm{gaze}}^{\mathrm{deg}}), \quad
    \hat{\mathbf{E}}_{\mathrm{trace}} = \tfrac{1}{2}(\mathbf{E}_{\mathrm{trace}}^{\mathrm{orig}} + \mathbf{E}_{\mathrm{trace}}^{\mathrm{deg}}),
\end{aligned}$}
\end{equation}
where superscripts $\mathrm{orig}$ and $\mathrm{deg}$ denote the original 
and degraded feature versions respectively, with degradation applied 
according to the modality degradation settings described in 
Sec.~\ref{sec:experiments}. The averaged embeddings 
$\hat{\mathbf{E}}_{\{\cdot\}} \in \mathbb{R}^{B \times d}$ provide stable 
representations irrespective of modality quality variations across 
students.

The stabilized embeddings are concatenated to form a joint multimodal 
representation:
\begin{equation}
    \mathbf{U}^{b} = [\hat{\mathbf{E}}_{\mathrm{AU}}, \hat{\mathbf{E}}_{\mathrm{pose}}, \hat{\mathbf{E}}_{\mathrm{gaze}}, \hat{\mathbf{E}}_{\mathrm{trace}}] \in \mathbb{R}^{B \times 4d},
\end{equation}
enabling direct semantic comparison across all modalities. Unlike 
cross-attention fusion that suffers from cascading error accumulation under 
modality degradation, this concatenation strategy treats all modalities 
equally and eliminates intermediate computations that amplify alignment 
errors.

To explicitly quantify semantic correlations among modalities, we compute 
an affinity matrix:
\begin{equation}
    \mathbf{A}^{b} = \sigma(\mathbf{U}^{b}(\mathbf{U}^{b})^{\top}),
\end{equation}
where $\sigma$ is a normalization function bounding affinity scores within 
$[0, 1]$. This symmetric positive semi-definite matrix captures explicit 
semantic relationships among AU, pose, gaze, and trace features, ensuring 
robust modality-level alignment.

Guided by the affinity matrix, we introduce a contrastive learning 
strategy to enforce semantic alignment and enhance cross-modal consistency. 
Within each batch, positive pairs are formed from embeddings of different 
modalities belonging to the same student activity, promoting cross-modal 
coherence. Negative pairs are constructed from embeddings across different 
student activities, enhancing discriminative power. The affinity-driven 
contrastive loss is formulated as:
\vspace{-2mm}
\begin{equation}
\scalebox{0.8}{$\displaystyle
\mathcal{L}_{\mathrm{aff}} = \frac{1}{B}\sum_{b=1}^{B} \left[\underbrace{\underset{i \neq j}{\sum^{m}_{b_1 = b}} (1 - \mathbf{A}_{i,j}^{b_1})^2}_{\text{push positive pairs}} + \underbrace{\underset{i \neq j}{\sum^{m}_{b_1 \neq b_2}} (\mathbf{A}_{i,j}^{b_1,b_2})^2}_{\text{pull negative pairs}}\right],$}
\end{equation}
where $\mathbf{A}_{i,j}^{b_1}$ represents the affinity score between 
modalities $i$ and $j$ for student activity $b_1$, and 
$\mathbf{A}_{i,j}^{b_1,b_2}$ denotes the cross-activity affinity score.

The proposed CAMA explicitly captures modality interactions and offers 
interpretable insights into cross-modal feature alignment across diverse 
student cohorts, even under modality degradation. Unlike prior implicit 
fusion approaches~\cite{li2021selfdoc}, CAMA employs explicit affinity 
matrices to model inter-modal relationships, mitigating the semantic 
inconsistencies that arise when gaze or other modalities carry inconsistent 
signals across students.

\vspace{-2mm}
\subsection{Training Objectives}

The total loss integrates the cross-entropy classification loss 
$\mathcal{L}_{\mathrm{CE}}$ with the affinity alignment loss 
$\mathcal{L}_{\mathrm{aff}}$:
\begin{equation}
    \mathcal{L}_{\mathrm{total}} = \mathcal{L}_{\mathrm{CE}} + \alpha \mathcal{L}_{\mathrm{aff}},
\end{equation}
where $\alpha$ is empirically set to 0.25 via grid search. This 
formulation ensures that \textbf{classification accuracy and cross-modal feature 
alignment are jointly optimized}, leading to a more robust and 
generalizable collaboration satisfaction prediction framework.

\vspace{-2mm}\section{Experiments Settings and Results}
\label{sec:experiments}
\begin{figure*}[h]
    \centering
    \begin{tabular}{cc}
        \includegraphics[width=0.48\textwidth]{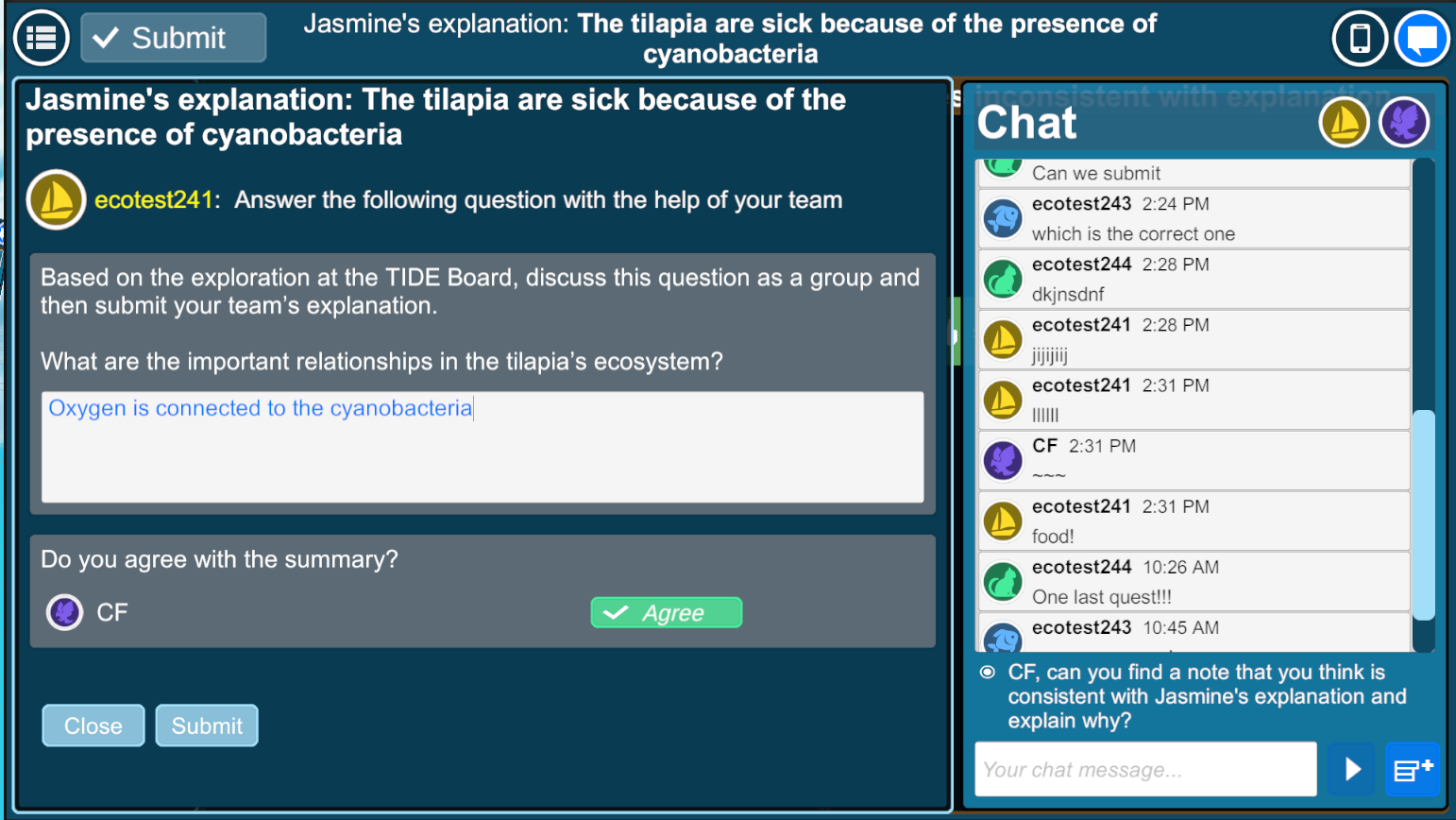} &
        \includegraphics[width=0.48\textwidth]{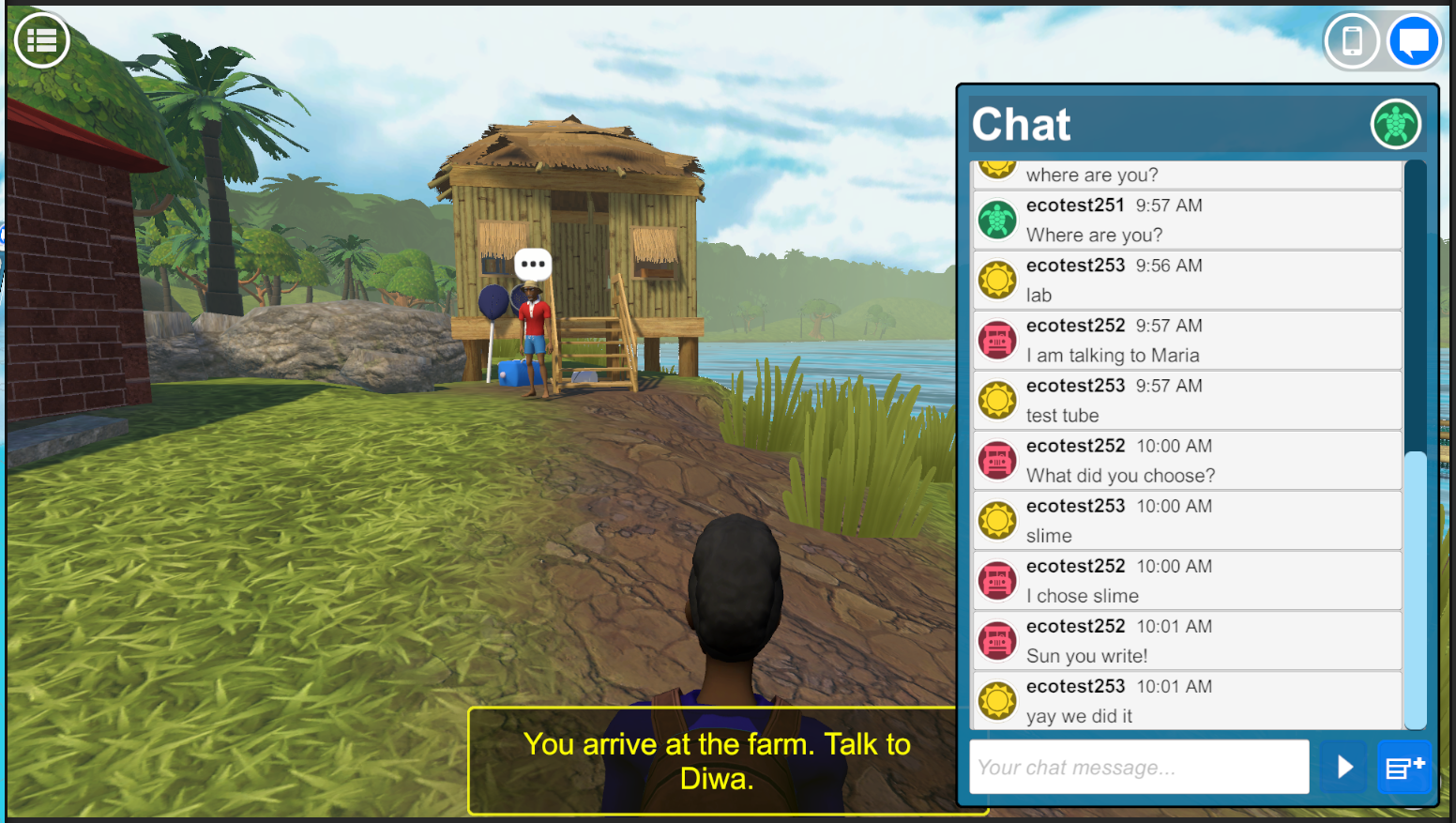} \\[4pt]
        \parbox{0.48\textwidth}{\centering (a) The TIDE Board interface, 
        where students collaboratively construct and submit explanations 
        alongside real-time group chat, generating rich trace log data.} &
        \parbox{0.5\textwidth}{\centering (b) The 3D exploration view, 
        where students navigate the virtual island, interact with NPCs 
        such as local farmers, and coordinate actions through the in-game 
        chat interface.}
    \end{tabular}\caption{\textbf{Student communication while playing in the EcoJourneys 
    collaborative learning environment~\cite{EDM19}.} Students work in 
    small groups to investigate a fish illness on a virtual Philippine 
    island, generating rich multimodal behavioral signals — including 
    facial expressions, head pose, eye gaze, and in-game chat interactions 
    — that our AAMLA framework leverages for collaboration satisfaction 
    prediction. Unlike prior implicit fusion approaches, AAMLA explicitly 
    aligns these heterogeneous modalities to suppress uninformative signals 
    and produce robust cross-modal representations.}
    \label{fig:ecojourneys}
    \vspace{-3mm}
\end{figure*}

\vspace{-2mm}
\noindent\textbf{Dataset.}
We utilize multimodal data collected from 50 middle school students 
(6th--8th grade, ages 11--14) interacting with the EcoJourneys 
collaborative game-based learning environment~\cite{carpenter2020detecting}, 
comprising 164 completed activities with paired video and trace log 
recordings. EcoJourneys is a problem-based collaborative learning 
environment in which small groups of students investigate the cause of an 
unknown illness affecting a fish population on a fictional Philippine island. 
As shown in Fig.~\ref{fig:ecojourneys}, students are guided by the TIDE 
inquiry cycle (Talk, Investigate, Deduce, Explain): they converse with 
non-player characters (NPCs) to gather domain knowledge 
(Fig.~\ref{fig:ecojourneys}b), explore the virtual island to collect 
evidence, and engage in collaborative reasoning via an in-game chat 
interface and a shared virtual whiteboard (Fig.~\ref{fig:ecojourneys}a). 
Students progress through four activities (one tutorial and three quests), 
with an exit survey administered after each activity to capture their 
sentiments regarding the collaborative experience.

Video data was captured via front-facing laptop cameras during gameplay, 
while trace logs documented fine-grained in-game actions including NPC 
interactions, locations visited, evidence collected, and chat messages 
exchanged. Collaboration satisfaction labels were derived from the 
Likert-scale exit surveys, categorized into four classes based on group 
listening and idea-building responses. We follow a student-level 10-fold 
cross-validation protocol to ensure that all activities from the same 
student appear exclusively in either training or testing splits.

\noindent\textbf{Feature Extraction and Synchronization.}
Video-based features are extracted using OpenFace 2.0~\cite{baltrusaitis2018openface} 
from front-facing camera recordings, yielding three complementary facial 
feature streams: (1) 17 AU intensity values per frame ranging from 0 
(absent) to 5 (high intensity); (2) 6 head pose features encoding 
3D translation and rotation coordinates; and (3) 6 gaze direction features 
encoding 3D coordinates per eye. Trace log events are encoded as 
768-dimensional embeddings using a BERT-based sentence encoder 
fine-tuned via unsupervised learning on in-game text~\cite{li2021selfdoc}, 
with a byte-pair encoding tokenizer to capture semantic nuances specific 
to in-game interactions.

Since facial features and trace log events are sampled at different rates, 
we follow~\cite{EDM19} and adopt trace log events as the base sampling 
rate, synchronizing facial features by averaging values between successive 
trace events. Formally, denoting facial features as $\mathbf{F}(t)$ and 
trace log events as $\mathbf{S}(t_k)$, where $t_k$ represents the slower 
trace sampling rate, synchronized facial features $\bar{\mathbf{F}}(t_k)$ 
are computed as:
\begin{equation}
    \bar{\mathbf{F}}(t_k) = \frac{1}{|\mathcal{I}_k|} \sum_{t \in \mathcal{I}_k} \mathbf{F}(t),
\end{equation}
where $\mathcal{I}_k$ denotes the interval $(t_{k-1}, t_k]$ and 
$|\mathcal{I}_k|$ is the count of facial feature frames within that 
interval. This yields a synchronized dataset wherein each trace event 
$\mathbf{S}(t_k)$ corresponds to a mean facial feature vector 
$\bar{\mathbf{F}}(t_k)$, forming the multimodal input sequences fed into 
our modality-specific encoders.

\noindent\textbf{Modality Degradation Settings.}
To systematically evaluate robustness against modality degradation --- a 
core motivation of our work --- we simulate three degradation conditions 
following prior work on missing and corrupted 
modalities~\cite{ma2021smil,wang2023multi,QAD}:
\begin{itemize}
    \item \textbf{Gaze dropout}: randomly zeroing gaze features at rates 
    of 30\%, 50\%, and 70\% of timesteps, simulating occlusion or 
    tracking failure.
    \item \textbf{Gaussian perturbation}: adding noise 
    $\mathcal{N}(0, \sigma)$ with $\sigma \in \{0.01, 0.05\}$ to AU 
    and pose features, simulating sensor noise and compression artifacts.
    \item \textbf{Full modality absence}: completely removing one modality 
    at inference, evaluating graceful degradation without retraining.
\end{itemize}
These settings reflect realistic challenges in educational deployments, 
where behavioral variation and hardware limitations cause modality quality 
to fluctuate across students and sessions.

\noindent\textbf{Hyperparameter Settings and Implementation Details.}
Models are trained with a batch size of 16 using the Adam 
optimizer~\cite{kingma2017adammethodstochasticoptimization} with an 
initial learning rate of $1 \times 10^{-3}$, reduced on validation loss 
plateau. Early stopping with patience of 3 is applied for regularization, 
with a maximum of 500 epochs. The affinity alignment weight $\alpha$ is 
set to 0.25, selected via grid search over $\{0.1, 0.25, 0.5\}$. Dropout 
of 0.1 is applied to all GRU encoders. All modality-specific features are 
projected into a unified $d=128$ dimensional space via learnable linear 
projection layers prior to affinity-driven alignment. Data synchronization 
follows the original protocol~\cite{carpenter2020detecting}, averaging facial features between successive trace events. Evaluation metrics 
include macro F1-score and accuracy, consistent with the class-imbalanced 
nature of the satisfaction labels. All experiments are implemented in 
PyTorch and run on a single NVIDIA A100 GPU.

\subsection{Unimodal Baseline Results}

We first evaluate the predictive performance of each modality 
independently using GRU-based unimodal models, establishing baseline 
benchmarks consistent with prior work~\cite{EDM19}. As shown in 
Table~\ref{tab:unimodal}, all unimodal models achieve significant 
improvement over a majority classifier, demonstrating that each modality 
captures meaningful cues for collaboration satisfaction prediction. AU and 
pose features yield the strongest unimodal performance (F1 = 0.66, Acc = 
0.66), while gaze and trace features perform comparably (F1 = 0.65, Acc = 
0.65). Wilcoxon signed-rank tests across cross-validation folds reveal no 
significant performance differences among most unimodal models, with the 
exception of pose versus gaze ($p < .05$), suggesting that these two 
modalities capture complementary but asymmetric information about 
collaborative dynamics.

\begin{table}[h]
    \centering
\caption{Comparison of unimodal baselines, cross-attention~\cite{EDM19}, 
    and AAMLA. Mean $\pm$ std. over 10 runs of student-level 10-fold 
    cross-validation.}
    \scalebox{0.75}{
    \begin{tabular}{lcc}
        \toprule
        Model & F1-Score & Accuracy \\
        \midrule
        AU (Unimodal)    & 0.66 {\color{teal}\small$\pm$0.03} & 0.66 {\color{teal}\small$\pm$0.03} \\
        Pose (Unimodal)  & 0.66 {\color{teal}\small$\pm$0.04} & 0.66 {\color{teal}\small$\pm$0.04} \\
        Gaze (Unimodal)  & 0.65 {\color{teal}\small$\pm$0.05} & 0.65 {\color{teal}\small$\pm$0.05} \\
        Trace (Unimodal) & 0.65 {\color{teal}\small$\pm$0.04} & 0.65 {\color{teal}\small$\pm$0.04} \\
        \midrule
        Cross-Attention~\cite{EDM19} & 0.72 {\color{teal}\small$\pm$0.03} & 0.72 {\color{teal}\small$\pm$0.03} \\
        \midrule
        \textbf{AAMLA (Ours)} & \textbf{0.79} {\color{teal}\small$\pm$0.02} & \textbf{0.77} {\color{teal}\small$\pm$0.02} \\
        \bottomrule
    \end{tabular}
    }
    \label{tab:unimodal}
\end{table}
\subsection{Multimodal Ablation Study}

To understand the contribution of each component in our framework, we 
conduct a comprehensive ablation study comparing AAMLA against the prior 
cross-attention baseline~\cite{EDM19} and intermediate model variants. 
Results are reported in Table~\ref{tab:ablation}.

\begin{table}[h]
    \centering
    \caption{Ablation study comparing model components. Proj.: modality 
    projection layers. CAMA: Cross-modal Affinity-guided Modality Alignment. 
    $\mathcal{L}_{\mathrm{aff}}$: contrastive alignment loss.}
    \scalebox{0.7}{
    \begin{tabular}{lcccccc}
        \toprule
        Model & Proj. & CAMA & $\mathcal{L}_{\mathrm{aff}}$ & F1 & Acc \\
        \midrule
        A: Cross-Attention~\cite{EDM19} 
        & \ding{55} & \ding{55} & \ding{55} & 0.72 & 0.72 \\
        B: + Projection 
        & \ding{51} & \ding{55} & \ding{55} & 0.71 & 0.72 \\
        C: + Projection + CAMA 
        & \ding{51} & \ding{51} & \ding{55} & 0.75 & 0.74 \\
        D: + Projection + CAMA + $\mathcal{L}_{\mathrm{aff}}$ (Full) 
        & \ding{51} & \ding{51} & \ding{51} & 0.79 & 0.77 \\
        \bottomrule
    \end{tabular}
    }
    \label{tab:ablation}
\end{table}
Model A reproduces the prior cross-attention baseline~\cite{EDM19}, 
serving as the primary point of comparison. Model B isolates the effect of 
projection-based feature space unification, examining whether aligning 
heterogeneous modality dimensions into a common $d=128$ space alone 
improves performance. Model C adds the CAMA affinity matrix fusion without 
contrastive supervision, assessing the contribution of explicit inter-modal 
relationship modeling. Model D represents the full AAMLA framework, 
incorporating all three components.

We additionally reproduce the modality combination ablation 
from~\cite{EDM19} under our framework to examine whether CAMA changes which 
modality combinations are most effective, as reported in 
Table~\ref{tab:modality_ablation}.
\begin{table}[h]
    \centering
    \caption{Modality combination ablation under AAMLA framework, compared 
    to prior cross-attention results~\cite{EDM19}.}
    \scalebox{0.75}{
    \begin{tabular}{lcccc}
        \toprule
        \multirow{2}{*}{Modality Combination} 
        & \multicolumn{2}{c}{Cross-Attn~\cite{EDM19}} 
        & \multicolumn{2}{c}{AAMLA (Ours)} \\
        \cmidrule(lr){2-3} \cmidrule(lr){4-5}
        & F1 & Acc & F1 & Acc \\
        \midrule
        Full Multimodal  & 0.72 & 0.72 & 0.79 & 0.77 \\
        Trace + AU       & 0.70 & 0.70 & 0.73 & 0.74 \\
        Trace + Pose     & 0.69 & 0.69 & 0.72 & 0.74 \\
        Trace + Gaze     & 0.65 & 0.65 & 0.70 & 0.71 \\
        AU + Pose        & 0.68 & 0.67 & 0.71 & 0.72 \\
        AU + Gaze        & 0.65 & 0.65 & 0.70 & 0.71 \\
        Pose + Gaze      & 0.64 & 0.64 & 0.69 & 0.68\\
        Trace + AU + Pose& 0.70 & 0.70 & 0.74 & 0.76 \\
        \bottomrule
    \end{tabular}
    }
    \label{tab:modality_ablation}
\end{table}

\vspace{-3mm}
\subsection{Modality Robustness Evaluation}

A critical advantage of AAMLA over the prior cross-attention 
approach~\cite{EDM19} is its explicit mechanism for handling uninformative 
or degraded modalities. We systematically evaluate robustness under the 
three degradation conditions defined in Sec.~\ref{sec:experiments}, 
comparing AAMLA against the cross-attention baseline~\cite{EDM19}. Results 
are summarized in Table~\ref{tab:robustness}.

\begin{table}[h]
    \centering
    \caption{\textbf{Modality robustness evaluation} comparing 
    Cross-Attention~\cite{EDM19} and AAMLA (Ours) under three degradation 
    conditions. Results reported as macro F1-score.}
    \scalebox{0.8}{
    \begin{tabular}{llcc}
        \toprule
        Modality & Degradation 
        & Cross-Attn~\cite{EDM19} & AAMLA (Ours) \\
        \midrule
        \multirow{3}{*}{Gaze} 
        & Dropout 30\%           & 0.68 & 0.77 \\
        & Dropout 50\%           & 0.61 & 0.75 \\
        & Dropout 70\%           & 0.52 & 0.72 \\
        \midrule
        \multirow{2}{*}{AU + Pose} 
        & $\mathcal{N}(0, 0.01)$ & 0.65 & 0.78 \\
        & $\mathcal{N}(0, 0.05)$ & 0.54 & 0.72 \\
        \midrule
        \multirow{4}{*}{Full Absence} 
        & w/o AU    & 0.54 & 0.73 \\
        & w/o Pose  & 0.53 & 0.72 \\
        & w/o Gaze  & 0.52 & 0.68 \\
        & w/o Trace & 0.56 & 0.70 \\
        \bottomrule
    \end{tabular}
    }
    \label{tab:robustness}
\end{table}

As demonstrated in the deepfake detection domain~\cite{umcl}, explicit 
affinity-based alignment creates inherent redundancy among modalities such 
that when one modality degrades, learned cross-modal relationships allow 
the remaining modalities to compensate. We hypothesize that AAMLA exhibits 
analogous behavior in the educational domain, particularly for gaze dropout 
conditions where the prior cross-attention model~\cite{EDM19} is known to 
be sensitive.

\begin{figure*}[t]
\begin{center}
\scalebox{0.5}{
    \renewcommand{\arraystretch}{0.4}
    \begin{tabular}[c]{c@{ }c@{ }c@{ }c@{ }}
        \hspace*{-8mm}\includegraphics[width=0.45\textwidth]{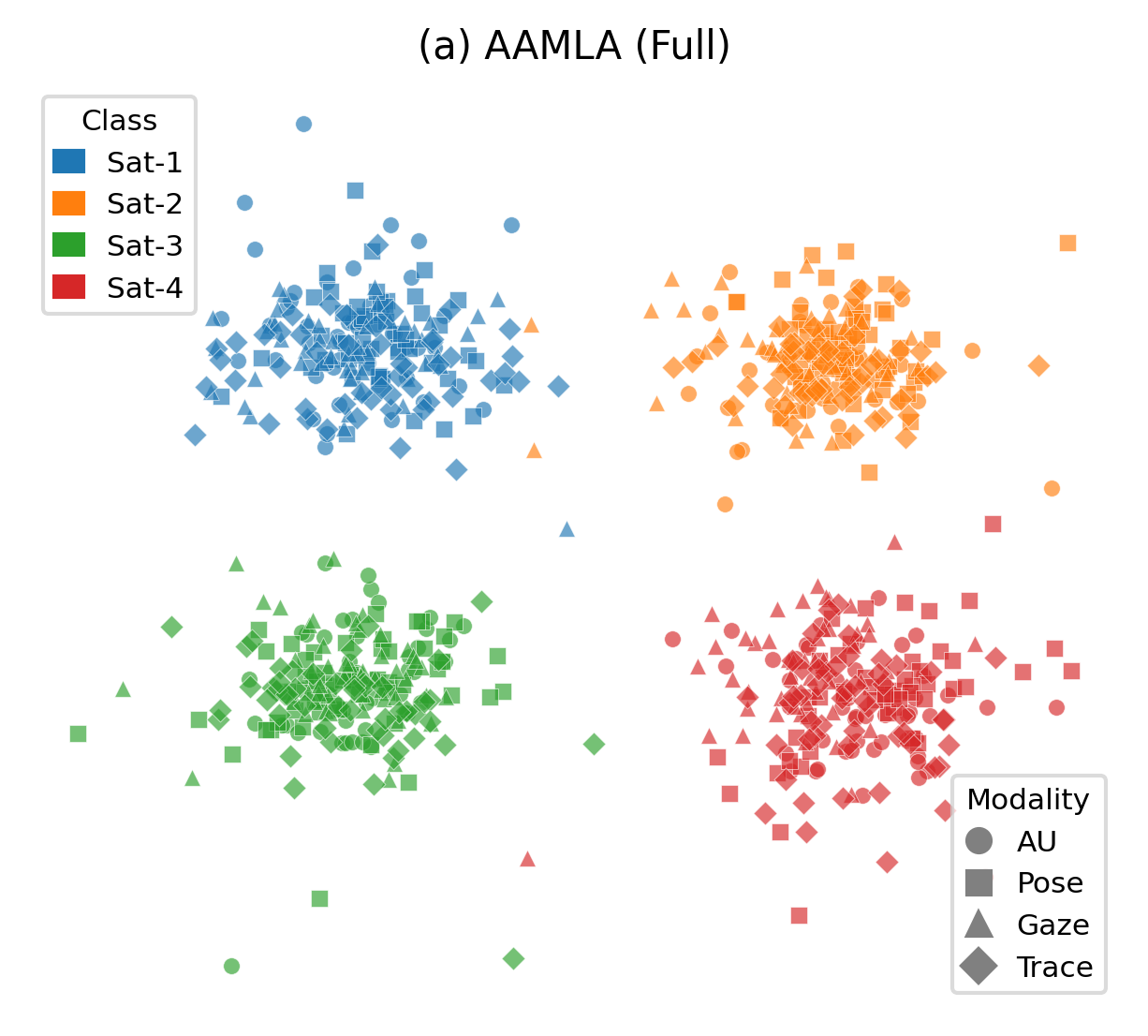}&
        \includegraphics[width=0.45\textwidth]{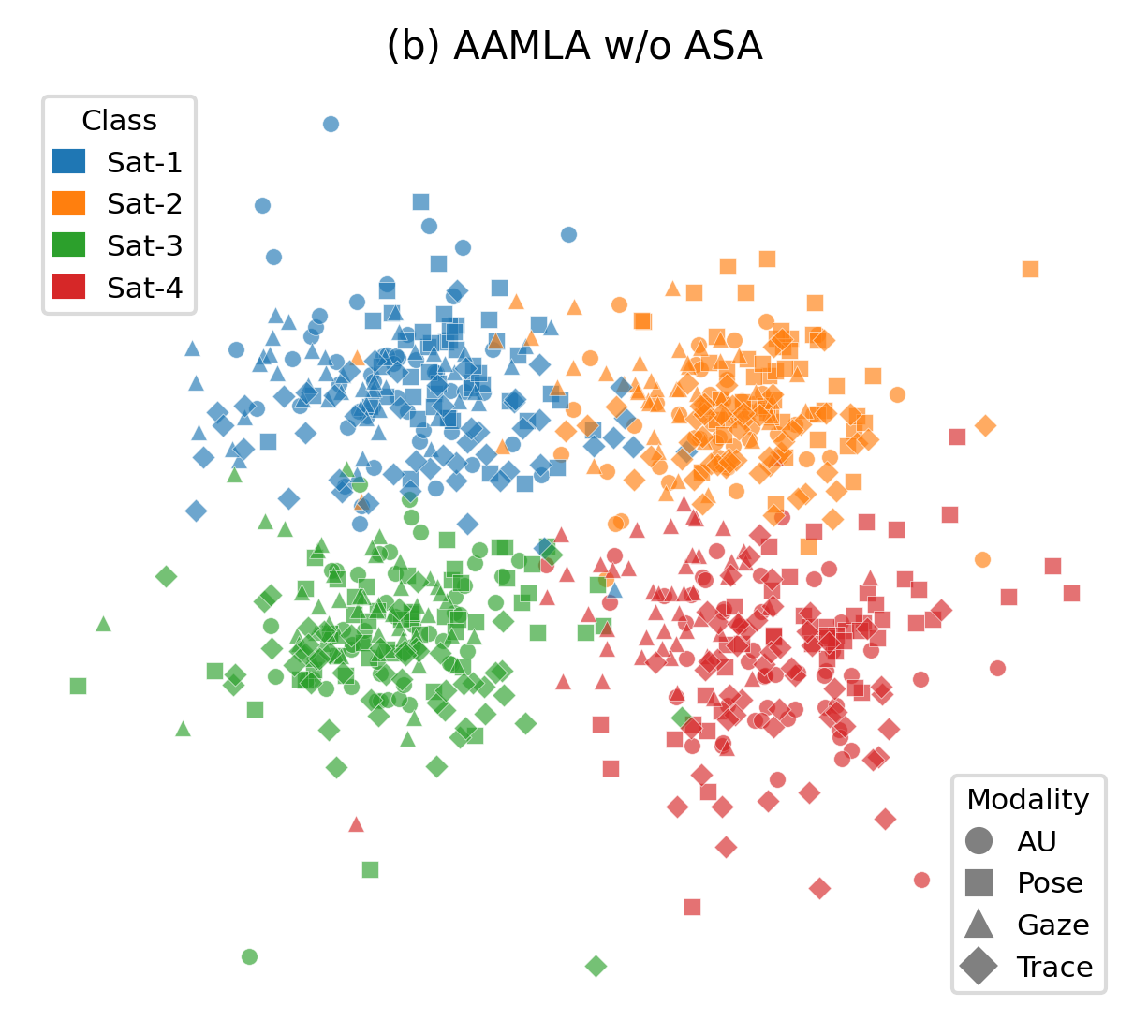}&
        \includegraphics[width=0.45\textwidth]{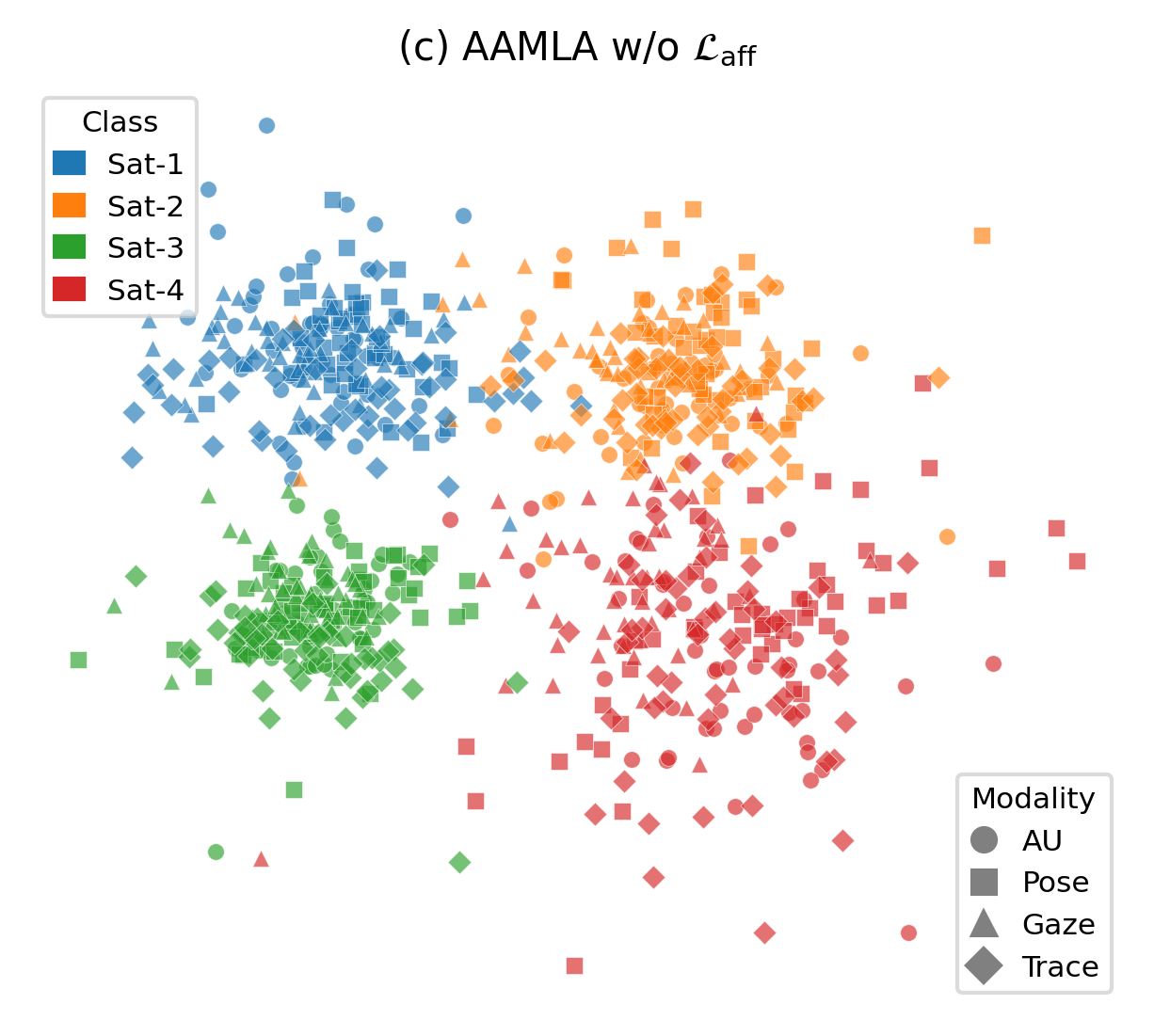}&
        \includegraphics[width=0.45\textwidth]{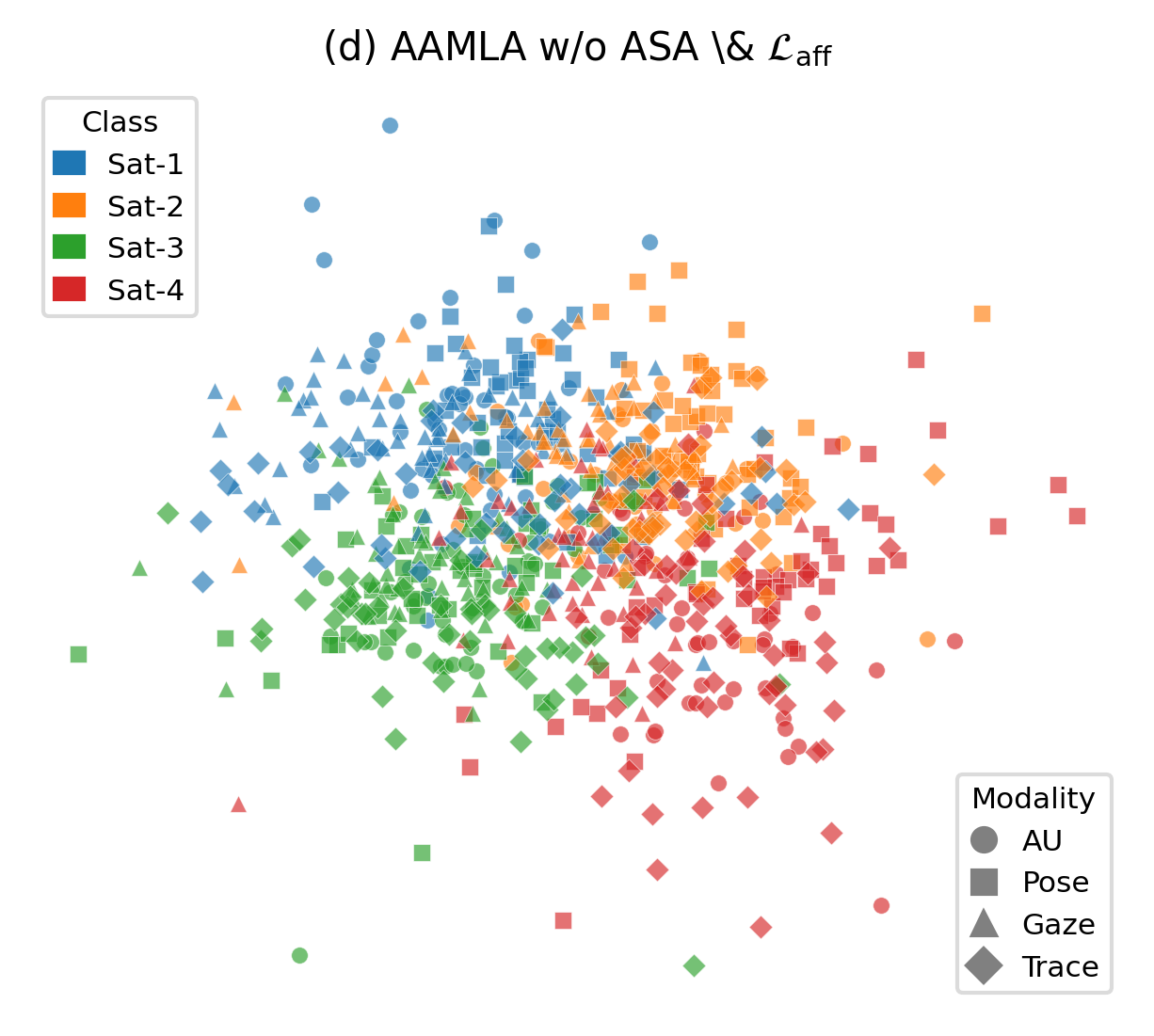}
        \\[4pt]
        \hspace*{-8mm}\Large~\parbox{0.45\textwidth}{\centering (a) AAMLA (Full)} &
        \Large~\parbox{0.45\textwidth}{\centering (b) AAMLA w/o CAMA} &
        \Large~\parbox{0.45\textwidth}{\centering (c) AAMLA w/o $\mathcal{L}_\mathrm{aff}$} &
        \Large~\parbox{0.45\textwidth}{\centering (d) AAMLA w/o CAMA \& $\mathcal{L}_\mathrm{aff}$}
    \end{tabular}
}
\end{center}
\vspace{-3mm}
\caption{\textbf{t-SNE visualizations of multimodal feature distributions 
under different ablation settings.} Color denotes satisfaction class 
(Sat-1 to Sat-4); marker shape denotes modality (AU, Pose, Gaze, Trace). 
(a) The full AAMLA model produces tightly clustered, semantically aligned 
features with clear inter-class separation. (b) Removing CAMA causes 
cross-modality drift and partial overlap between satisfaction classes. 
(c) Removing $\mathcal{L}_\mathrm{aff}$ produces fragmented decision 
boundaries with increased cohort-level variance. (d) Removing both 
components yields the most scattered feature space, with blurred 
inter-class boundaries and widened inter-modality semantic gaps.}
\label{fig:tsne}
\end{figure*}

\begin{figure*}[t]
\begin{center}
\scalebox{0.5}{
    \renewcommand{\arraystretch}{0.4}
    \begin{tabular}[c]{c@{ }c@{ }c@{ }c@{ }}
        \hspace*{-8mm}\includegraphics[width=0.45\textwidth]{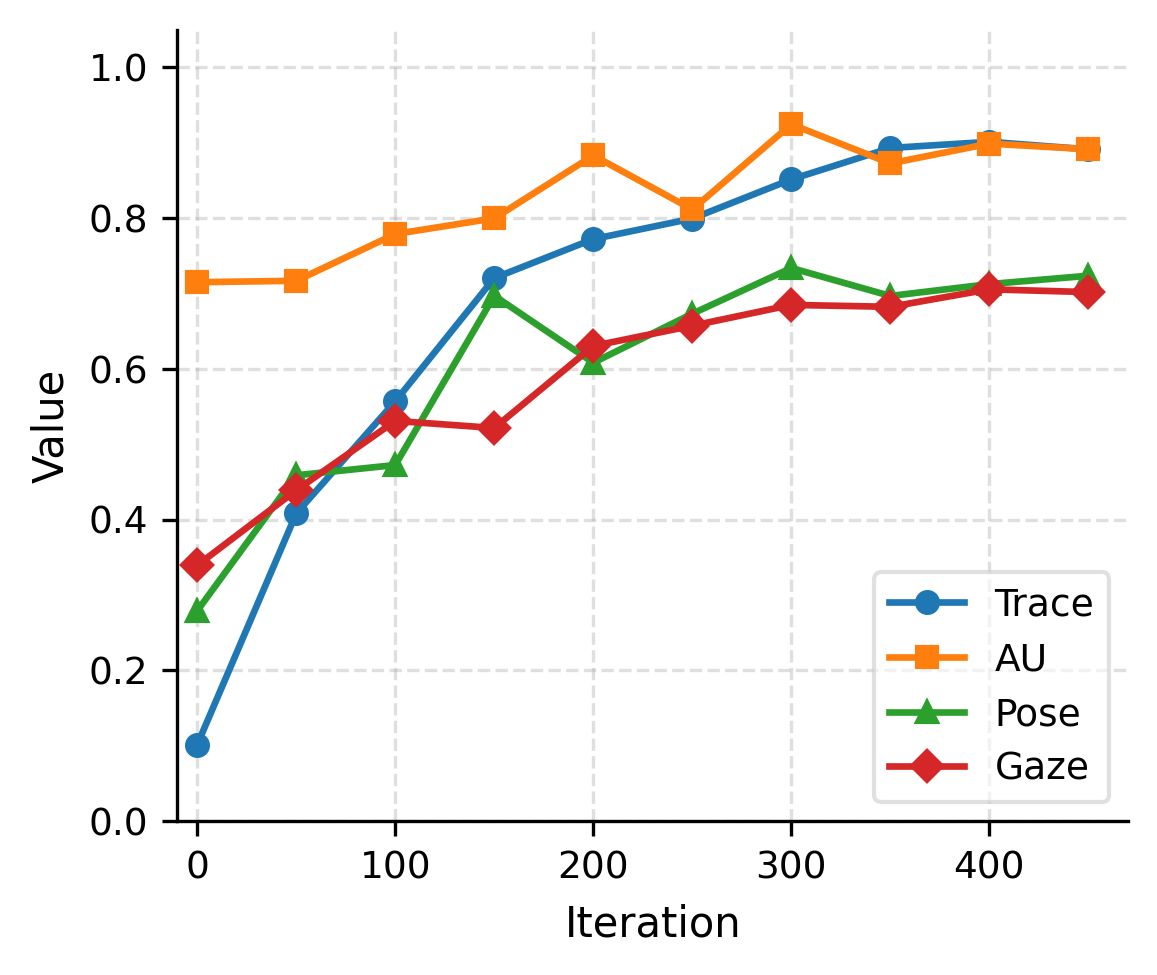}&
        \includegraphics[width=0.45\textwidth]{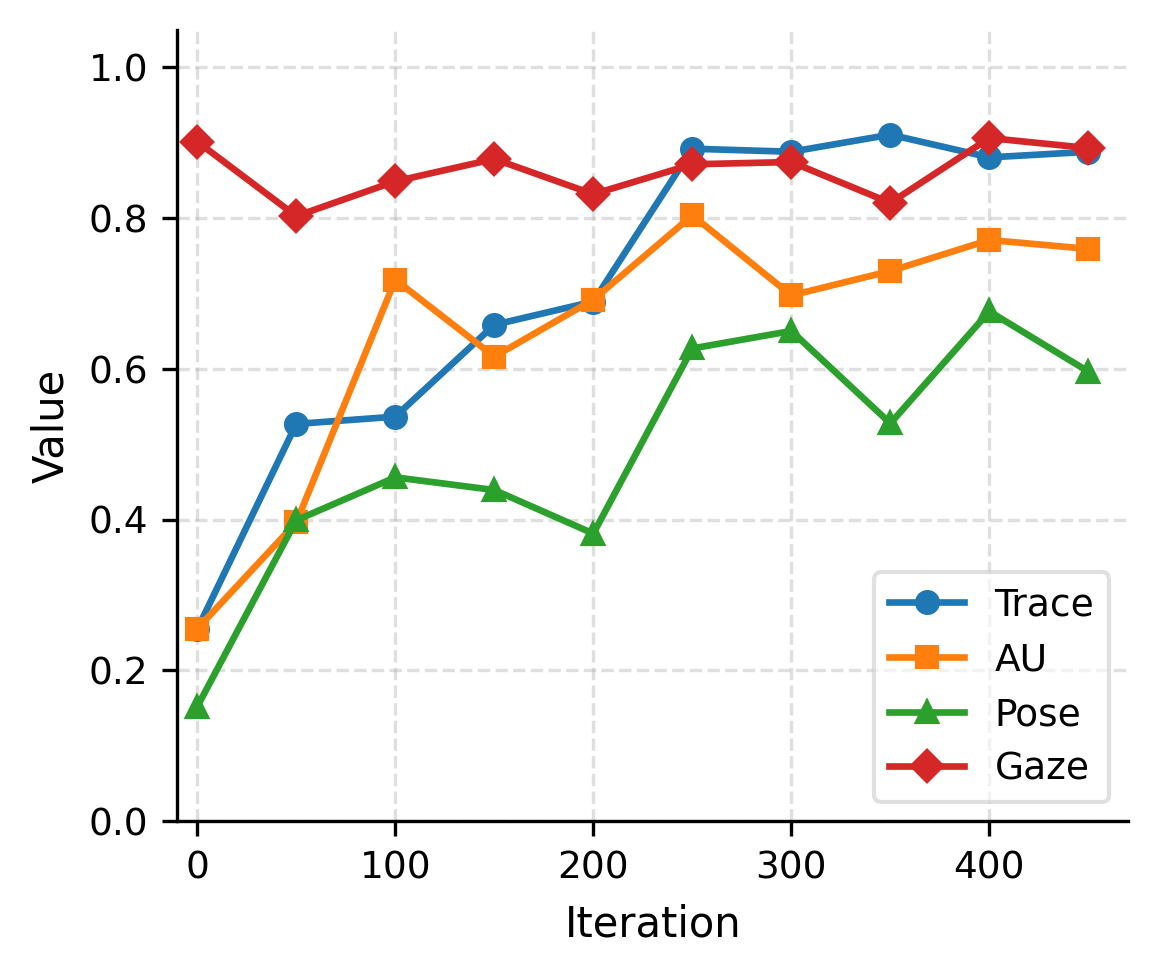}&
        \includegraphics[width=0.45\textwidth]{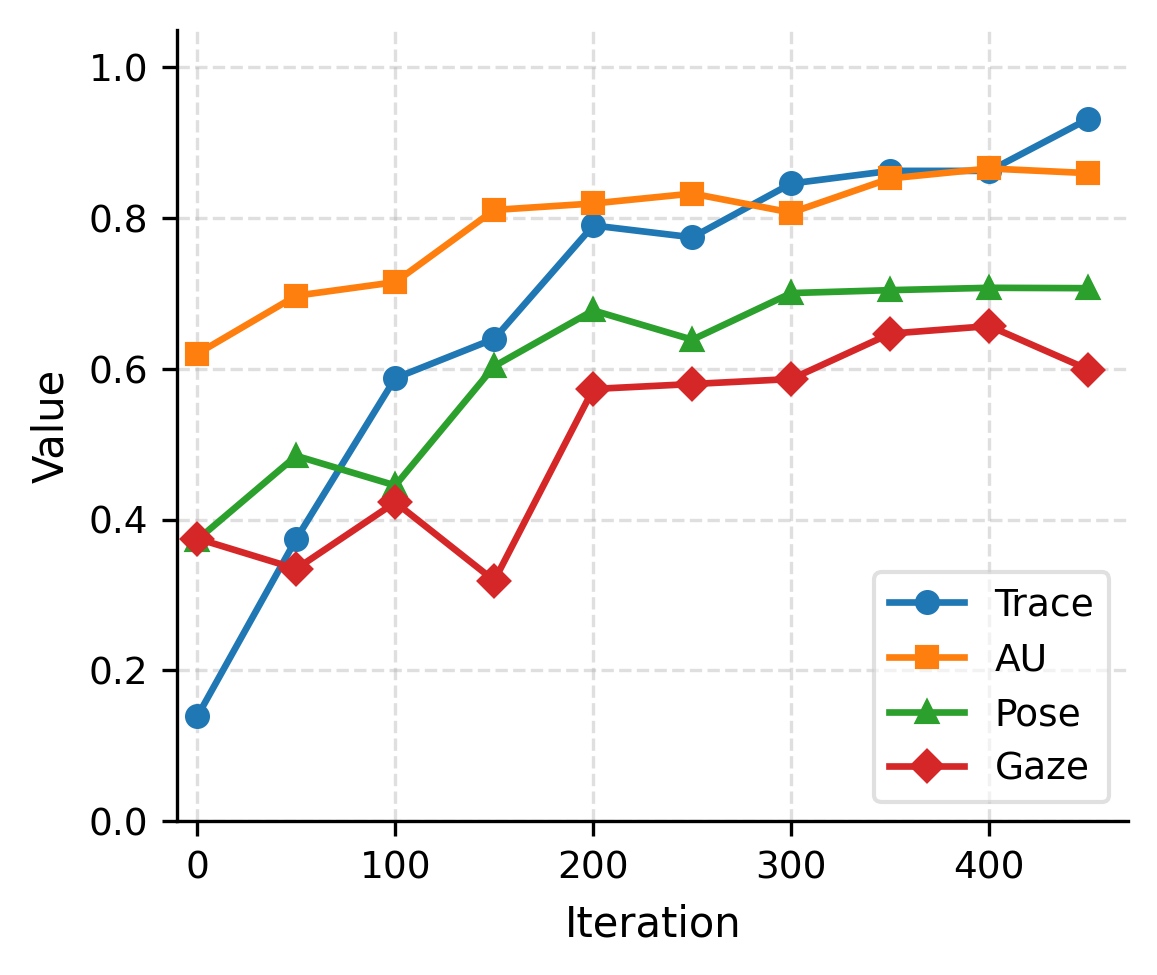}&
        \includegraphics[width=0.45\textwidth]{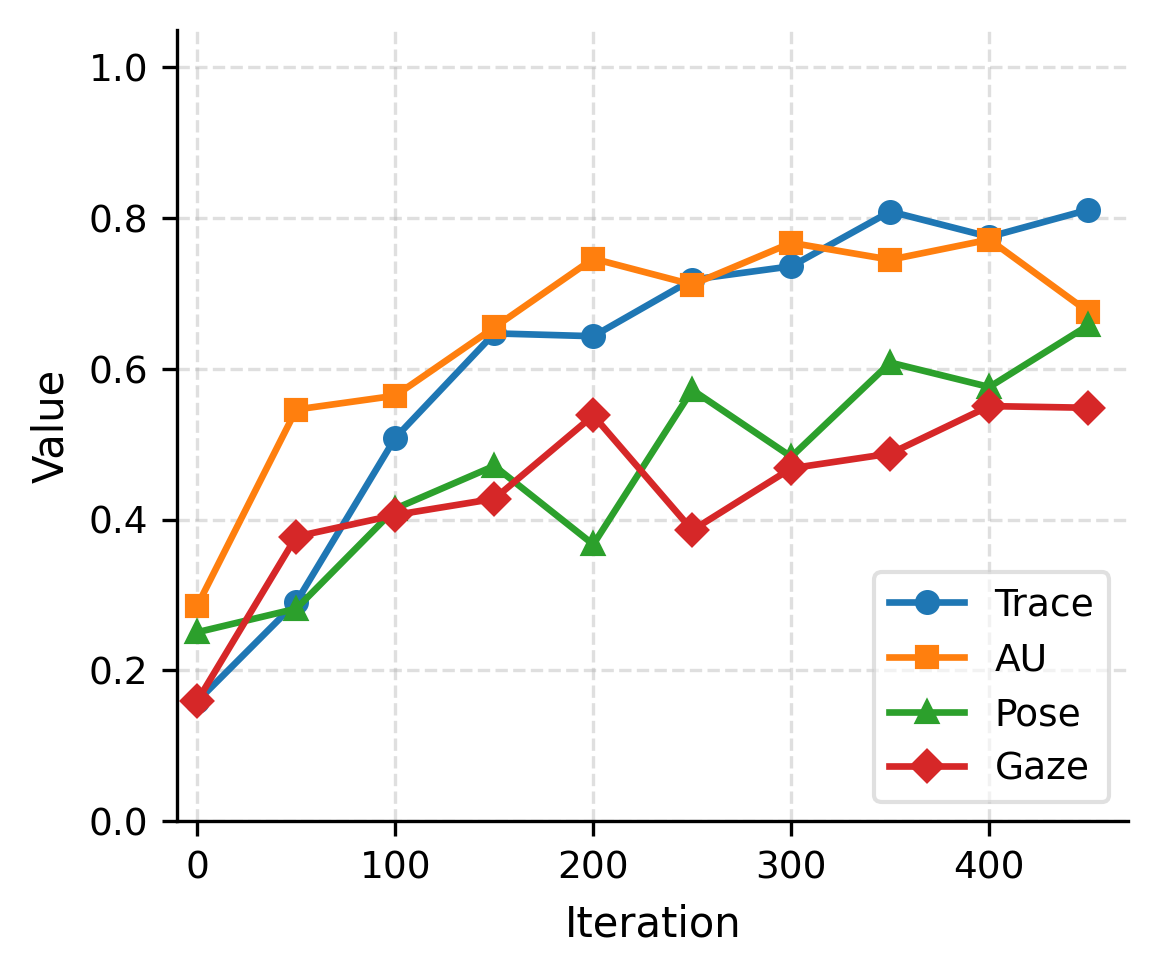}
        \\[4pt]
        \hspace*{-8mm}\Large~\parbox{0.45\textwidth}{\centering (a) High-satisfaction\\affinity scores} &
        \Large~\parbox{0.45\textwidth}{\centering (b) Low-satisfaction\\affinity scores} &
        \Large~\parbox{0.45\textwidth}{\centering (c) Degraded high-satisfaction\\affinity scores} &
        \Large~\parbox{0.45\textwidth}{\centering (d) Degraded low-satisfaction\\affinity scores}
    \end{tabular}
}
\vspace{-1mm}
\end{center}
\vspace{-3mm}
\caption{\textbf{Affinity scores evolution among AU, pose, gaze, and trace modalities 
across $\{\text{high-satisfaction},\text{low-satisfaction}\}$ and 
$\{\text{original},\text{degraded}\}$ conditions during training.} 
High-satisfaction activities achieve stable convergence earlier, 
reflecting more consistent cross-modal alignment, while degraded 
conditions exhibit higher variance, particularly for gaze features,
motivating the explicit alignment enforced by CAMA.}
\label{fig:CAMA_vis}
\vspace{-3mm}
\end{figure*}

\subsection{Analyses}

\noindent\textbf{t-SNE Feature Distribution.}
Fig.~\ref{fig:tsne} presents t-SNE visualizations under four ablation 
settings. The full AAMLA model [Fig.~\ref{fig:tsne}(a)] produces 
well-separated clusters across the four satisfaction classes, with 
modality-specific embeddings remaining semantically cohesive. Removing CAMA 
[Fig.~\ref{fig:tsne}(b)] introduces visible cross-modality drift and 
inter-class overlap, reflecting weaker inter-modal consistency. Without 
$\mathcal{L}_{\mathrm{aff}}$ [Fig.~\ref{fig:tsne}(c)], cluster boundaries 
become fragmented with uneven variance across cohorts. The fully unaligned 
model [Fig.~\ref{fig:tsne}(d)] yields the most scattered feature space, 
where class boundaries blur and inter-modality gaps widen substantially. 
Together, these results confirm that CAMA and $\mathcal{L}_{\mathrm{aff}}$ 
jointly contribute to compact, discriminative representations.

\noindent\textbf{Affinity Score Evolution.}
Fig.~\ref{fig:CAMA_vis} shows affinity score trajectories during training. 
Trace embeddings converge most stably, consistent with their semantic 
richness as a behavioral modality, while AU and pose exhibit more dynamic 
patterns reflecting their interaction dependencies. Gaze features show the 
lowest and most variable scores, corroborating prior observations of their 
inconsistent informativeness across student cohorts~\cite{EDM19}. 
High-satisfaction activities reach stable alignment earlier than 
low-satisfaction ones, suggesting that positive collaborative interactions 
produce more consistent cross-modal patterns.

\begin{figure}[h]
    \centering
    \includegraphics[width=0.41\textwidth]{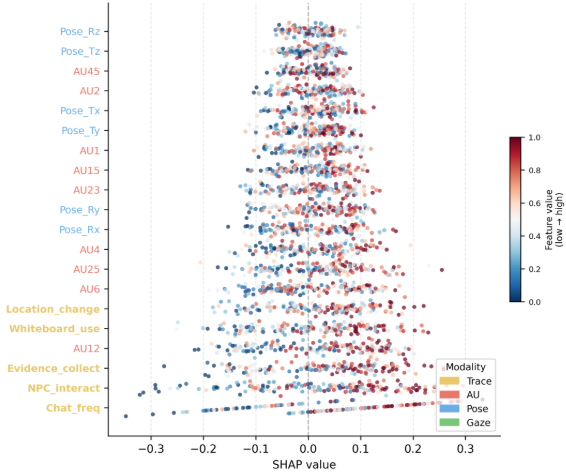}
    \caption{\textbf{SHAP beeswarm plot of feature contributions}. 
    Color denotes normalized feature value. 
    Trace features rank highest, gaze features are absent from the 
    top-20, corroborating CAMA's adaptive suppression of uninformative 
    modalities.}
    \vspace{-2mm}
    \label{fig:shap}
\end{figure}

\noindent\textbf{SHAP Analysis.}
As shown in Fig.~\ref{fig:shap}, trace log features (\texttt{Chat\_freq}, 
\texttt{NPC\_Interact}, \texttt{Evidence\_collect}) consistently rank 
highest, confirming their dominance in satisfaction prediction. AU and 
pose features show moderate contributions, while gaze features are absent 
from the top-20 — directly validating CAMA's adaptive suppression of 
uninformative modalities.

\vspace{-2mm}\section{Conclusion}
\label{sec:conclusion}

\vspace{-2mm}
In this paper, we presented AAMLA, an Affinity-Aligned Multimodal Learning Analytics 
framework that addresses modality degradation in collaborative game-based 
learning through Cross-modal Affinity-guided Modality Alignment (CAMA). 
By explicitly modeling inter-modal relationships via affinity matrices and 
contrastive learning, CAMA adaptively suppresses uninformative modalities 
such as gaze without discarding them, producing robust representations 
that generalize across diverse student cohorts. Experiments on EcoJourneys 
confirm consistent improvements over unimodal baselines and the prior 
cross-attention approach under both standard and degraded conditions, with 
SHAP and t-SNE analyses providing interpretable insights into the 
behavioral signals most predictive of collaboration satisfaction. More 
broadly, this work demonstrates that explicit multimodal alignment 
techniques transfer effectively to educational settings where modality 
reliability is inherently variable.

{
    \small
    \bibliographystyle{ieeenat_fullname}
    \bibliography{main}
}

\end{document}